\documentclass[10pt]{article} 
\usepackage[preprint]{tmlr}

\usepackage{hyperref}
\usepackage{url}
\usepackage{booktabs}
\usepackage{graphicx}
\usepackage{amsmath}
\usepackage{amssymb}
\usepackage{microtype}
\usepackage{xcolor}
\usepackage{multirow}
\usepackage{array}
\usepackage{float}
\usepackage{placeins}

\title{When Chain-of-Thought Helps and When It Hurts:\\
An Empirical Investigation of the\\
Serial-Depth Bottleneck in LLM Reasoning}

\author{\name Tughanbulut Kurtulush \email tkurtul1@stu.vistula.edu.pl \\
      \addr Faculty of Computer Engineering \\
      Vistula University, Warsaw, Poland \\
      ORCID: 0009-0009-4861-7126}

\def\month{05}
\def\year{2026}

\begin{document}

\maketitle

\begin{abstract}
It is widely assumed that chain-of-thought (CoT) prompting universally
improves LLM reasoning. We investigate this assumption through the
conceptual framework of the $H_{\!dp}$ bandwidth bound
\citep{chen2024theoreticallimitationsmultilayertransformer}. While the
formal bound applies only asymptotically -- at astronomically large
prompt lengths -- it identifies a fundamental architectural bottleneck:
serial computation whose depth exceeds a transformer's
single-forward-pass capacity must be externalised, precisely what CoT
does. Our central empirical finding is a within-benchmark serial-depth
gradient: single-pass (no-CoT) accuracy degrades monotonically with
per-item serial depth, while CoT is approximately depth-invariant. We measure CoT effects
across three instruction-tuned models (Qwen-2.5-7B/32B, Llama-3.1-8B)
and five standard NLP benchmarks at practical context lengths. The
depth-class hypotheses inspired by the framework are otherwise
supported. On high-depth P-complete tasks (GSM8K, MATH), CoT provides a massive $+54$
to $+68$ percentage-point recovery gap across all three models.
Conversely, on shallow TC$^0$ tasks (MMLU, ARC-Challenge), forcing
CoT reasoning is
structurally redundant: it yields approximately zero benefit
($\Delta \in [0.0, +4.6]$\,pp across all six cells, no
Bonferroni-significant negative effect); these high TC$^0$ baselines
(up to 95\% on ARC) may, however, reflect pretraining contamination rather than
genuinely shallow computation, so this null is not a clean architectural test. Tasks in the intermediate class~$\mathbf{L}$
(HumanEval) exhibit a strict model-size-dependent transition: $+23.2$\,pp
for the 32B model, $+9.1$\,pp for the 8B, $-28.7$\,pp for the 7B. The
pooled cross-benchmark depth--recovery correlation is Spearman $\rho = 0.661$
($p = 0.007$, $n = 15$), with 9 of 15 benchmark-level McNemar tests
significant after Bonferroni correction. Our findings, pre-registered
on OSF, indicate that chain-of-thought is not a universal reasoning
enhancer but acts as a \textbf{bandwidth bypass} -- helping serial
computation that strains single-pass capacity while remaining redundant
for tasks that already fit.
\end{abstract}

\section{Introduction}

A central question in LLM theory is \emph{which} computational problems a
transformer can solve within a single forward pass.
\citet{chen2024theoreticallimitationsmultilayertransformer} formalise
this via the \emph{multi-party autoregressive communication model}.
Their main result (Theorem~1.1) is the first \emph{unconditional} lower bound for
multi-layer decoder-only transformers: an $L$-layer transformer with $H$ attention
heads of head dimension $d$ and precision $p$ cannot solve $L$-sequential function
composition whenever $H_{\!dp} = H \times d \times p \leq n^{2^{-4L}}$, where $n$
is the prompt length.
Equivalently, $H_{\!dp}$ is the key bandwidth parameter governing how much serial
computation a single forward pass can perform.
As a direct corollary, the paper provides a \emph{provable advantage of
chain-of-thought}: tasks that are exponentially hard for a single forward pass
become exponentially easier once intermediate steps are externalised to the
output stream. Figure~\ref{fig:theorem} (in Section 2) illustrates this
contrast schematically.

Despite its theoretical elegance, the $H_{\!dp}$ bound describes an asymptotic
regime. Inverting its condition, the smallest prompt length at which the failure
guarantee is non-vacuous for a given model is $n^\star = H_{\!dp}^{\,2^{4L}}$,
which for the models studied here ranges from $n^\star \approx 10^{10^{34.4}}$
(Qwen-7B) to $n^\star \approx 10^{10^{77.8}}$ (Qwen-32B) -- far beyond any
physically realizable sequence length. It is therefore unknown whether this
architectural bottleneck has any bearing on the benchmarks routinely used to
compare deployed LLMs, whose prompts span only a few hundred to a few thousand
tokens.

\textbf{This paper tests that gap.}
We (1) map five widely-used NLP benchmarks onto the CC primitives from
\citet{chen2024theoreticallimitationsmultilayertransformer}, (2) derive
qualitative, depth-class hypotheses motivated by the $H_{\!dp}$ bound for each benchmark,
and (3) test whether two conditions -- direct-answer (no-CoT) versus
chain-of-thought (CoT) -- produce the accuracy patterns the serial-depth account
implies.

The pre-registered hypothesis had two parts. On the positive side: high-depth
(P-complete) benchmarks should show large CoT recovery gains because their high
serial depth should exceed single-pass capacity. On the negative side: low-depth (TC$^0$)
benchmarks should be CoT-insensitive or actively penalised, because additional
serial tokens add noise without unlocking computation the model lacks
bandwidth for. We find the positive hypothesis supported across all three
models and both math benchmarks. The negative hypothesis, with correctly
extracted answers (see Appendix~\ref{sec:correction}), is not supported on MMLU or
ARC, where CoT is approximately neutral across all six cells
($\Delta \in [0.0, +4.6]$\,pp). The only \emph{negative} cell that survives
correct extraction is HumanEval/Qwen-7B ($-28.7$\,pp), which is a class~$\mathbf{L}$
benchmark rather than a TC$^0$ one. These findings converge with the
large-scale meta-analysis of \citet{sprague2025cot}, who show across 100+
papers and 20 datasets that CoT helps mainly on math and symbolic reasoning,
with little to no benefit on non-symbolic tasks.

\section{Background}

\subsection{The $H_{\!dp}$ Bound}

\begin{figure}[htbp]
    \centering
    \includegraphics[width=0.8\textwidth]{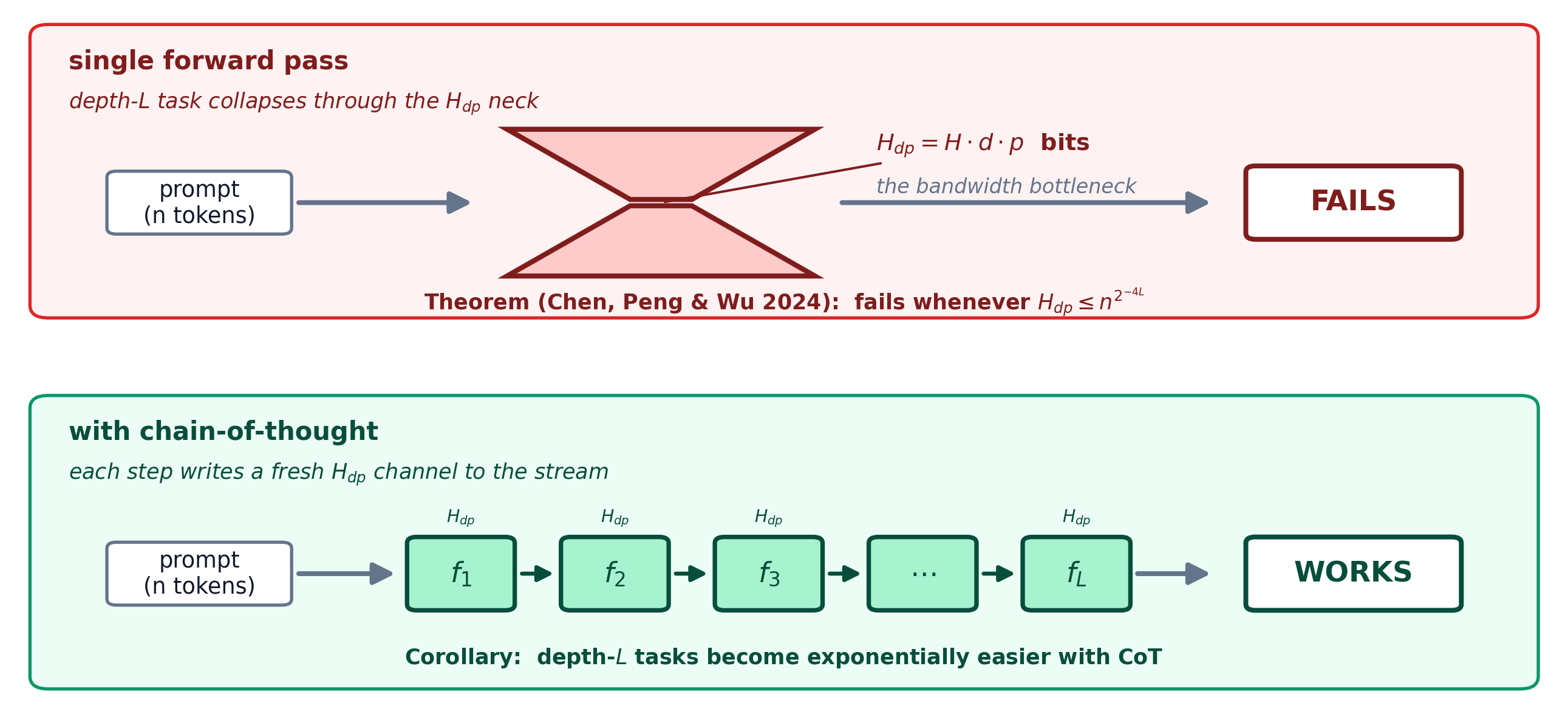}
    \caption{Schematic of the Chen, Peng \& Wu (2024) theorem, included as
    conceptual motivation. \textbf{Top}: in the single-forward-pass regime a
    depth-$L$ task is bottlenecked at the $H_{dp} = H \times d \times p$ bit neck,
    and the theorem guarantees failure whenever $H_{dp} \leq n^{2^{-4L}}$.
    \textbf{Bottom}: chain-of-thought externalises each step to the output stream,
    giving each layer a fresh $H_{dp}$ channel and making depth-$L$ tasks
    exponentially easier. The bound is asymptotic and binds only at astronomically
    large $n$, far beyond the context lengths we test; the figure illustrates the
    theoretical mechanism, not a claim about empirical behaviour.}
    \label{fig:theorem}
\end{figure}

Let a decoder-only transformer have $L$ layers, $H$ attention heads per layer,
head dimension $d$, and numerical precision $p$ bits (16 for FP16).
The \emph{bandwidth parameter} is:
\[
    H_{\!dp} \;=\; H \times d \times p.
\]
\citet{chen2024theoreticallimitationsmultilayertransformer} prove
unconditionally that an $L$-layer decoder-only transformer fails at
$L$-sequential function composition whenever $H_{\!dp} \leq n^{2^{-4L}}$.
Crucially, this is an asymptotic bound. Inverting the condition, the smallest
prompt length at which the guarantee is non-vacuous for a model of bandwidth
$H_{\!dp}$ is $n^\star = H_{\!dp}^{\,2^{4L}}$.\footnote{Setting the threshold
equal to the model's bandwidth, $n^{2^{-4L}} = H_{\!dp}$, and solving gives
$n^\star = H_{\!dp}^{\,2^{4L}}$.} For the models tested this ranges from
$n^\star \approx 10^{10^{34.4}}$ (Qwen-7B, $L{=}28$) to
$n^\star \approx 10^{10^{77.8}}$ (Qwen-32B, $L{=}64$) -- versus the
$n \approx 10^2$--$10^3$ tokens of our benchmarks. The formal bound therefore
does not bind at any physically realizable context length, and we treat it as a
conceptual motivation rather than a literal predictor. It nonetheless identifies
a real architectural property: decoder-only transformers possess a serial-depth
bottleneck, and tasks requiring more serial depth than the single-forward-pass
bandwidth permits must externalise intermediate computation -- precisely what
chain-of-thought does. We hypothesise that this bottleneck governs empirical
reasoning failures even at practical sequence lengths, making it the mechanistic
motivation for the CoT recovery gains we measure on P-complete benchmarks.

This architectural constraint was anticipated empirically by
\citet{nye2021scratchpad}, who noted that a model
\textit{``is asked to perform these tasks in one forward pass [and] cannot
adapt the amount of compute''}, and showed that routing intermediate steps
through an explicit scratchpad recovers accuracy on multi-step arithmetic
that fails under direct prediction.
The $H_{\!dp}$ bound offers a conceptual explanation for why the scratchpad
helps precisely on tasks requiring serial depth.

\subsection{CC Primitives and Benchmark Mapping}
\label{sec:mapping}

We assign each benchmark a coarse heuristic mapping into the CC-primitive
taxonomy of
\citet{chen2024theoreticallimitationsmultilayertransformer}, intended
solely as a vehicle for deriving pre-registered, depth-class hypotheses:

\begin{table}[ht]
\centering
\small
\begin{tabular}{llll}
\toprule
\textbf{Benchmark} & \textbf{Heuristic primitive} & \textbf{Depth class} & \textbf{CoT pred.} \\
\midrule
GSM8K~\citep{cobbe2021trainingverifierssolvemath}   & $k$-seq. composition  & P-complete & $+$ \\
MATH~\citep{hendrycks2021measuringmathematicalproblemsolving}  & Nested $k$-comp.      & P-complete & $+$ \\
MMLU~\citep{hendrycks2021measuring}  & Set disjointness      & TC$^0$     & $-$ / $0$ \\
ARC-C~\citep{clark2018thinksolvedquestionanswering}      & Sparse parity         & TC$^0$     & $-$ / $0$ \\
HumanEval~\citep{chen2021evaluatinglargelanguagemodels}    & Pointer chasing       & $\mathbf{L}$ & moderate \\
\bottomrule
\end{tabular}
\caption{Benchmark-to-primitive mapping (\emph{heuristic}). The
``Heuristic primitive'' column is the initial single-primitive label
used to derive pre-registered hypotheses. An LLM-as-judge inter-rater
check (\S\ref{sec:results}) yields pooled $\kappa = 0.293$, indicating
that real benchmarks are mixtures of primitives rather than instances of
any one. The load-bearing column for the hypotheses tested
(H1--H4) is \emph{Depth class}, not the specific primitive label; the
depth-class hypotheses are robust to primitive relabeling within the
same class.}
\label{tab:mapping}
\end{table}

$k$-sequential composition (GSM8K, MATH) requires chaining $k$ arithmetic
operations where each depends on the previous -- a P-complete task.
\citet{cobbe2021trainingverifierssolvemath} define GSM8K problems
as requiring 2--8 sequential steps, directly quantifying the $k$ range.
Critically, they also provide their own empirical confirmation of the serial-depth
bottleneck: finetuning a 6B model to output the final answer \emph{without}
intermediate steps drops accuracy from 20.6\% to 5.2\% -- a 75\% relative
collapse caused purely by removing the intermediate computation steps.
This is precisely the no-CoT vs.\ CoT contrast we study, and their result
directly foreshadows our finding that P-complete benchmarks degrade sharply under
single-pass generation.
Set disjointness and sparse parity (MMLU, ARC) require only constant-depth
circuits and are in TC$^0$; they benefit little from serial scratchpad use.
MMLU~\citep{hendrycks2021measuring} spans 57 subjects from elementary mathematics
to professional law, evaluated zero-shot or few-shot from pretraining knowledge alone.
We provisionally treat MMLU as predominantly set-disjointness-like
(TC$^0$): most questions require selecting the answer option whose
content overlaps with a fact stored in the model, with no serial
composition between options. This is a coarse aggregate assignment; the
per-item picture is reported in \S\ref{sec:results}.
This classification is supported within the MMLU paper itself: Hendrycks et al.\
find that GPT-3 performs worst on calculation-heavy STEM subjects (which require
sequential arithmetic -- a P-complete operation) and best on verbal knowledge subjects
(factual retrieval -- TC$^0$).
The aggregate MMLU signal is therefore dominated by the TC$^0$ majority of subjects,
implying no CoT benefit.
This expectation is independently confirmed by \citet{sprague2025cot},
who show that 95\% of MMLU's total CoT performance gain is attributable to
questions whose text or model response contains an equals sign -- i.e., the
math-related minority -- while non-math MMLU questions receive no reliable
benefit from CoT.
\citet{clark2018thinksolvedquestionanswering} constructed the ARC
Challenge partition to contain \emph{only} questions that both a
retrieval-based solver and a word-co-occurrence solver answer
incorrectly; in 2018, no baseline -- including neural models trained
on SQuAD and SNLI -- significantly outperformed a random baseline
($\approx 25\%$) on the Challenge Set. Our models achieve 82--95\%
accuracy under no-CoT on the same partition. We do not interpret this
gap as evidence for any particular cause: modern instruction-tuned
models genuinely are far more capable than 2018 baselines, and
pretraining exposure to ARC is also plausible given the benchmark's
ubiquity. We return to this in \S\ref{sec:discussion}.
HumanEval~\citep{chen2021evaluatinglargelanguagemodels} evaluates Python
function synthesis from docstrings. We provisionally label the dominant
primitive as \emph{pointer chasing} -- the model must resolve a chain
of variable bindings and function calls in order. The per-item judge
(\S\ref{sec:results}) more often labels HumanEval items as
$k$-composition; both pointer chasing and $k$-composition involve
serial dependency, and pointer chasing is computable in log-space
(class~$\mathbf{L}$), placing the benchmark intermediate between
TC$^0$ and P-complete. We hypothesise a moderate, model-size-dependent CoT
benefit on this class.
This classification is corroborated by \citet{chen2021evaluatinglargelanguagemodels} themselves:
they report that Codex accuracy on synthetic chaining tasks degrades by a factor
of 2--3 per additional operation and that the model fails to bind operations to
variables correctly as chain length grows -- the signature of a serial-depth
bottleneck on sequential computation.

\paragraph{What is load-bearing.}
Throughout the remainder of this paper, the load-bearing column of
Table~\ref{tab:mapping} is \textbf{Depth class} (P-complete, TC$^0$,
class $\mathbf{L}$), not the specific primitive label. The primitive
labels are heuristic, per-item-noisy, and -- as the LLM-as-judge check
in \S\ref{sec:results} shows ($\kappa = 0.293$) -- not stable across
graders. The depth classes are coarse, hold up to primitive
relabeling within the same class, and are what hypotheses H1--H4
are derived from.

\section{Experimental Setup}

\subsection{Models}

We evaluate three instruction-tuned decoder-only transformer models spanning
a $1.43\times$ range of $H_{\!dp}$ bandwidth (Table~\ref{tab:models}).

\begin{table}[H]
\centering
\small
\begin{tabular}{lrrrr}
\toprule
\textbf{Model} & $L$ & $H$ & $d$ & $H_{\!dp}$ \\
\midrule
Qwen-2.5-7B-Instruct~\citep{qwen25}  & 28 & 28 & 128 & 57{,}344 \\
Llama-3.1-8B-Instruct~\citep{llama3} & 32 & 32 & 128 & 65{,}536 \\
Qwen-2.5-32B-Instruct~\citep{qwen25} & 64 & 40 & 128 & 81{,}920 \\
\bottomrule
\end{tabular}
\caption{Model architectures and $H_{\!dp}$ values ($p = 16$ for FP16).
$H$ = query heads; Llama-3.1-8B uses GQA with 8 KV heads, but output
bandwidth per token is determined by query heads × head dim. All models are
instruction-tuned variants without internal reasoning phases; the single-pass
condition we test applies directly to their single-forward-pass computation.}
\label{tab:models}
\end{table}

\FloatBarrier

\subsection{Conditions}

\textbf{no-CoT}: a direct-answer system prompt with a short output cap:
32 tokens for GSM8K, MMLU, and ARC; 64 for MATH (to allow
\texttt{\textbackslash boxed\{\}} formatting); and 256 for HumanEval (the
minimum required to emit a viable function body). All caps are
$\geq 8\times$ smaller than the CoT budget. For GSM8K, MATH, MMLU, and ARC the
short caps prevent the model from externalising multi-step reasoning to the
output stream, directly operationalising the single-forward-pass condition; the
256-token HumanEval cap is a necessarily looser operationalisation (a function
body cannot be written in a few tokens), a caveat we revisit in
\S\ref{sec:discussion}.

\textbf{CoT}: a standard chain-of-thought prompt~\citep{wei2022cot} with a
2048-token cap. Intermediate steps are externalised to the output stream,
bypassing the single-pass constraint.
This is mechanistically equivalent to the \emph{scratchpad} of
\citet{nye2021scratchpad}: both route intermediate state through
the output stream rather than requiring it to be compressed into residual
activations within a single forward pass.

\paragraph{Why the short-output cap denies externalised computation.}
We use ``single-pass condition'' as shorthand: generating $m$ output tokens is
strictly $m$ sequential forward passes, so the short caps do not enforce a
literal single forward pass. What they enforce is the absence of an
\emph{externalised scratchpad} -- the model cannot write intermediate state to
the output stream and read it back over many tokens, which is the operational
contrast the scratchpad/$H_{\!dp}$ account concerns. For instruction-tuned
models (which lack a hidden reasoning phase, unlike explicit reasoning models
such as DeepSeek-R1 or o1), the short caps deny this externalisation channel. A reader might object that the cap
simply provides insufficient output length to write the final answer. Three
observations rule this out. First, TC$^0$ tasks (MMLU, ARC) achieve 68--95\%
accuracy under the 32-token cap, demonstrating that 32 tokens is sufficient when
the computation fits in a single pass. Second, the cross-benchmark depth--recovery correlation (Spearman
$\rho = 0.661$ across 15 cells; \S\ref{sec:results}) shows the CoT gain scales
with assigned CC depth, not with output length. Third, the HumanEval
direction reversal between Qwen-7B ($-28.7$\,pp) and Qwen-32B ($+23.2$\,pp) holds the cap
fixed while varying model scale, so the effect is not an output-length artefact.
Reasoning models with explicit hidden chain-of-thought phases would require
a different experimental design and are outside our scope.

\subsection{Data, Inference, and Pre-registration}
\label{sec:prereg}

We sample 800 items sequentially from each benchmark's test split
(164 for HumanEval, full split).
All inference is run at temperature\,=\,0.0 (greedy decoding) in FP16 via
vLLM~\citep{kwon2023vllm} on rented GPU instances (Vast.ai).
Total compute cost: under \$5 USD.

\paragraph{Pre-registration.}
This study was formally pre-registered on the Open Science Framework (OSF
Registries) on 9~May 2026 under license CC-BY~4.0. The H1--H4 hypotheses
were derived from the $H_{\!dp}$ bound prior to any data collection;
production inference was partially complete at the time of registration
(Qwen-2.5-7B accuracy on GSM8K and MMLU had been observed, with no
stratified analysis performed), as disclosed in the registration's
foreknowledge statement. The registration is available at:
\begin{itemize}\setlength{\itemsep}{0pt}
  \item \textbf{Registration DOI:} \texttt{10.17605/OSF.IO/92JDK}
  \item \textbf{OSF page:} \url{https://osf.io/92jdk}
  \item \textbf{Associated project:} \url{https://osf.io/hteuj}
  \item \textbf{Internet Archive snapshot:}
        \url{https://archive.org/details/osf-registrations-92jdk-v1}
\end{itemize}
The pre-registration document specifies: (i) the four primary hypotheses
(H1--H4) tested in \S\ref{sec:results}; (ii) the McNemar exact test with
Bonferroni correction at $\alpha = 0.05/15$ as the primary statistical procedure;
(iii) the Spearman rank correlation as the secondary depth--recovery test;
(iv) the inter-rater Cohen's $\kappa \geq 0.7$ target for LLM-as-judge primitive
validation; and (v) the benchmark-to-primitive mapping used to derive hypotheses.
Of these pre-registered analyses, only the $\kappa$ target was not met; this
deviation is reported transparently in \S\ref{sec:results} and
\S\ref{sec:limitations}.

\subsection{Scoring}

GSM8K and MMLU/ARC use exact-match numeric and letter extraction respectively,
with regular-expression parsers robust to special-token leakage.
MATH uses a brace-balanced \texttt{\textbackslash boxed\{\}} extractor with
numeric equivalence for fractions.
HumanEval is functionally scored: the model output is parsed for the longest
fenced \texttt{```python} block (or, if absent, treated as a free-form
completion appended to the function prompt), then executed against the
canonical test suite in a 10\,s subprocess sandbox. All 984 HumanEval
inference records (164 items $\times$ 3 models $\times$ 2 conditions) are
scored; results match the pass@1 figures reported in Table~\ref{tab:results}.
All parsers were fixed and rows re-scored before any analysis;
Appendix~\ref{sec:correction} documents two scoring artefacts present in an
earlier preprint (Zenodo, May 2026) and their corrections.
Format error rate across all non-HumanEval scored inferences: 0.90\% under CoT
(86/9{,}600) and 0.46\% under no-CoT (44/9{,}600); excluded from accuracy
calculations but retained in the database.

\paragraph{Inter-rater agreement (Cohen's $\kappa$).}
The pre-registration committed to validating the per-benchmark single-primitive
labels via an independent LLM-as-judge. We used Gemma-2-27B-it
(\texttt{google/gemma-2-27b-it}, December 2024 release; family-independent of all
test models) at temperature 0, prompted with neutral CC-theory definitions and
required to output a one-sentence reasoning step plus a JSON
\texttt{(primitive, depth)} judgement for 200 sampled items per benchmark
(964 total; 0 parse failures).
For HumanEval the canonical implementation was provided alongside the function
spec so that the judge could observe the actual computational structure rather
than only the docstring.

\section{Results}
\label{sec:results}

\subsection{Main Effect}

\begin{figure}[ht]
    \centering
    \includegraphics[width=0.7\textwidth]{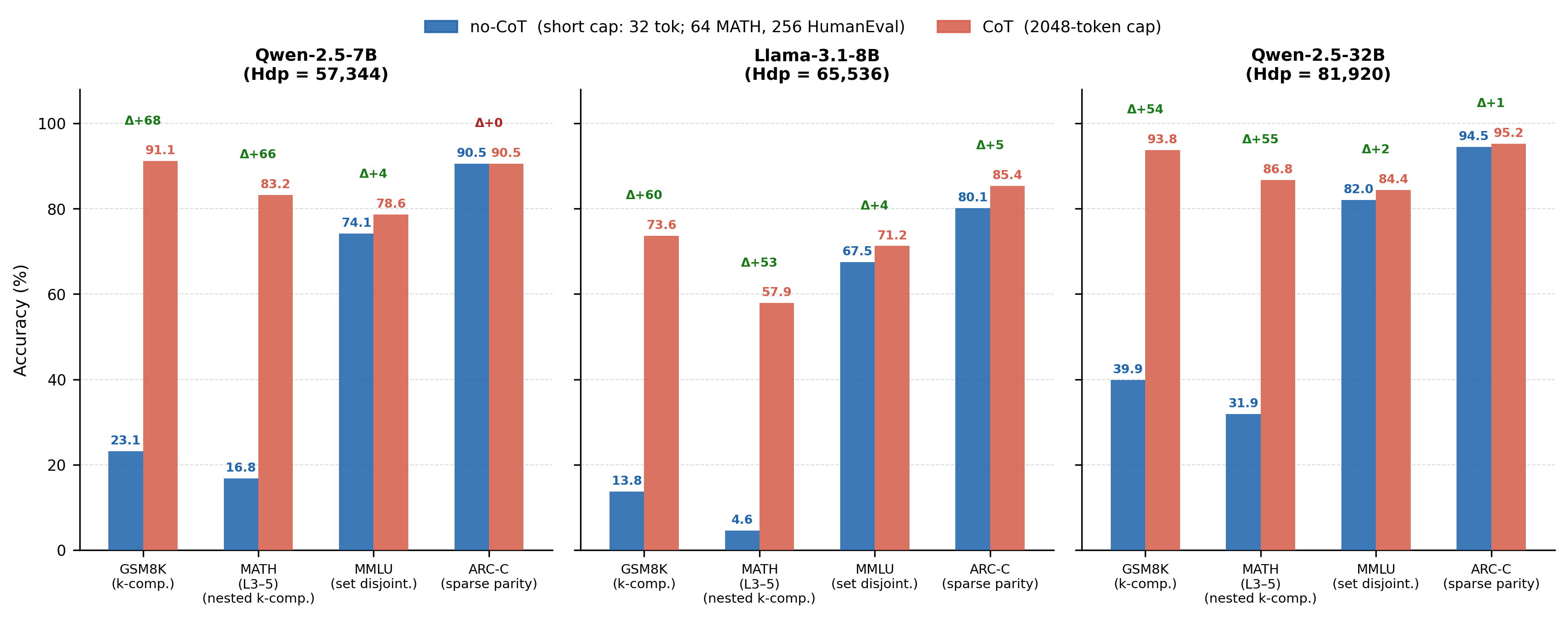}
    \caption{Accuracy by CoT condition and CC primitive. Green $\Delta$ = CoT
    gain; red $\Delta$ = CoT penalty. $n = 800$ per cell.}
    \label{fig:main}
\end{figure}

Figure~\ref{fig:main} shows accuracy under both conditions for all three models.
The math-side hypothesis holds across all three models: P-complete benchmarks
(GSM8K, MATH) show large positive CoT recovery gaps ($+54$ to $+68$\,pp). The
stronger negative TC$^0$ hypothesis does not hold: with correctly extracted
answers, CoT is approximately neutral on MMLU and ARC-Challenge across all
six (model, benchmark) cells ($\Delta \in [0.0, +4.6]$\,pp). HumanEval shows
the hypothesised model-size-dependent transition.

Table~\ref{tab:results} reports accuracy with 95\% Wilson confidence intervals.
The mean CoT recovery gap across models is $+60.6$\,pp for GSM8K,
$+60.5$\,pp for MATH, $+3.2$\,pp for MMLU, $+1.4$\,pp for ARC, and
$+1.2$\,pp for HumanEval (averaging $-28.7$, $+9.1$, and $+23.2$
across the three models).

\begin{table}[ht]
\centering
\footnotesize
\setlength{\tabcolsep}{5pt}
\renewcommand{\arraystretch}{0.88}
\begin{tabular}{llccc}
\toprule
\textbf{Model} & \textbf{Bench} & \textbf{no-CoT (\%)} & \textbf{CoT (\%)} & $\boldsymbol{\Delta}$ (pp) \\
\midrule
\multirow{5}{*}{Qwen-7B}
 & GSM8K & 23.1\scriptsize{ [20.3,26.2]} & 91.1\scriptsize{ [89.0,92.9]} & $+$68.0 \\
 & MATH  & 16.8\scriptsize{ [14.4,19.5]} & 84.3\scriptsize{ [81.6,86.7]} & $+$67.5 \\
 & MMLU  & 74.1\scriptsize{ [71.0,77.0]} & 78.7\scriptsize{ [75.8,81.4]} & $+$4.6 \\
 & ARC-C & 90.5\scriptsize{ [88.3,92.3]} & 90.5\scriptsize{ [88.3,92.3]} & $+$0.0 \\
 & HumanEval & 74.4\scriptsize{ [67.2,80.5]} & 45.7\scriptsize{ [38.3,53.4]} & $-$28.7 \\
\midrule
\multirow{5}{*}{Llama-8B}
 & GSM8K & 13.8\scriptsize{ [11.5,16.3]} & 73.6\scriptsize{ [70.5,76.6]} & $+$59.9 \\
 & MATH  &  4.7\scriptsize{ [3.4,6.4]}   & 63.3\scriptsize{ [59.7,66.7]} & $+$58.6 \\
 & MMLU  & 68.8\scriptsize{ [65.5,71.9]} & 71.3\scriptsize{ [68.1,74.4]} & $+$2.5 \\
 & ARC-C & 82.2\scriptsize{ [79.3,84.7]} & 85.5\scriptsize{ [82.9,87.8]} & $+$3.3 \\
 & HumanEval & 51.8\scriptsize{ [44.2,59.3]} & 61.0\scriptsize{ [53.3,68.1]} & $+$9.1 \\
\midrule
\multirow{5}{*}{Qwen-32B}
 & GSM8K & 39.9\scriptsize{ [36.5,43.3]} & 93.8\scriptsize{ [91.9,95.2]} & $+$53.9 \\
 & MATH  & 31.9\scriptsize{ [28.8,35.2]} & 87.3\scriptsize{ [84.8,89.4]} & $+$55.4 \\
 & MMLU  & 82.0\scriptsize{ [79.2,84.5]} & 84.4\scriptsize{ [81.7,86.7]} & $+$2.4 \\
 & ARC-C & 94.5\scriptsize{ [92.7,95.9]} & 95.2\scriptsize{ [93.5,96.5]} & $+$0.8 \\
 & HumanEval & 62.2\scriptsize{ [54.6,69.3]} & 85.4\scriptsize{ [79.1,90.0]} & $+$23.2 \\
\bottomrule
\end{tabular}
\caption{Accuracy (\%) with 95\% Wilson CIs [in brackets]. $\Delta$ = CoT $-$ no-CoT. $n=800$ per cell for GSM8K/MATH/MMLU/ARC; $n=164$ for HumanEval (full split).}
\label{tab:results}
\end{table}

\subsection{Statistical Tests}

\textbf{H1 (CoT effect; high-$k$ collapse).}
We first test, at the benchmark level, whether CoT changes accuracy at all:
McNemar's exact test, Bonferroni-corrected $\alpha = 0.05/15 = 0.0033$ for the
15 benchmark-level tests (5 benchmarks $\times$ 3 models). The depth-stratified
collapse that H1 concerns is tested at the bin level below.
\textbf{9 of 15 are significant.} The 9 significant cells are the six
P-complete cells (GSM8K and MATH across all three models, all with
$b \gg c$ and $p < 10^{-117}$), Qwen-7B's MMLU ($b{=}80$, $c{=}44$,
$p = 1.6 \times 10^{-3}$, in the direction CoT $>$ no-CoT), and two
HumanEval cells (Qwen-7B and Qwen-32B). The six \emph{non-significant} cells are
all on MMLU or ARC where the framework hypothesised no CoT benefit:
Qwen-7B/ARC ($b{=}39$, $c{=}39$, $p = 1.00$ -- a literal tie),
Llama-8B/MMLU ($b{=}84$, $c{=}61$, $p = 0.067$),
Llama-8B/ARC ($b{=}61$, $c{=}38$, $p = 0.027$),
Qwen-32B/MMLU ($b{=}46$, $c{=}27$, $p = 0.034$),
Qwen-32B/ARC ($b{=}20$, $c{=}14$, $p = 0.39$),
as well as Llama-8B/HumanEval ($b{=}25$, $c{=}10$, $p = 0.017$).
\textbf{Partially confirmed:} the math-side hypothesis holds across the
board; the TC$^0$ hypothesis (no CoT penalty) holds for all six MMLU/ARC
cells, but the framework's stronger claim that CoT should
\emph{actively hurt} TC$^0$ tasks is not supported (no cell shows a
significant negative direction at Bonferroni $\alpha$). The
HumanEval cells follow the hypothesised model-size-dependent pattern.

The pre-registration further specified 60 bin-level tests
(5 benchmarks $\times$ 3 models $\times$ 4 depth bins). With the updated
per-item depth labels (calculator-step counts for GSM8K, AST nesting depth for
HumanEval, equation-count proxy for MATH), 42 of these 60 cells are populated;
the remaining 18 are structurally empty because MMLU and ARC heuristic depth
labels collapse to $k{=}1$, consistent with their TC$^0$ classification.
Of the 42 computable bin-level tests, 25 are significant at the
Bonferroni-corrected $\alpha = 0.05/60 = 8.3 \times 10^{-4}$
(adjusted using the pre-registered denominator).
The non-significant cells are concentrated in HumanEval at $k{=}1$ and
$k{\geq}7$ (small bin counts: $n{=}1$ to $n{=}21$), in Llama-8B/HumanEval
where the CoT effect is small ($\Delta = +9.1$\,pp), and in
all five MMLU/ARC ($k{=}1$) cells that are non-significant at the
benchmark level above.

\textbf{H2 (recovery gap $\propto k$):}
Spearman rank correlation between assigned CC depth class
(P-complete benchmarks and HumanEval coded $k=4$,\footnote{HumanEval is coded at the P-complete depth ($k{=}4$) because the per-item judge predominantly labels its items as $k$-composition (\S\ref{sec:results}); this is the conservative choice, as it sets the strictest depth expectation for the class-$\mathbf{L}$ benchmark.} TC$^0$ coded $k=1$) and CoT
recovery gap, computed across all five benchmarks.
Pooled $\rho = 0.661$, $p = 0.007$ ($n=15$).
Per-model: Llama-8B $\rho = 0.866$ ($p = 0.058$, $n=5$),
Qwen-32B $\rho = 0.866$ ($p = 0.058$), Qwen-7B $\rho = 0.289$
($p = 0.638$) -- Qwen-7B's gradient is broken by its HumanEval
penalty ($-28.7$\,pp, hypothesised positive at depth $k{=}4$).
\textbf{Pooled gradient confirmed; per-model gradient holds for the two
larger models only.}

\textbf{H3 (MMLU: no-CoT $\geq$ CoT):}
One-sided McNemar tests $p = 1.00$, $0.98$, $0.99$ for Qwen-7B, Llama-8B,
Qwen-32B respectively -- all non-significant. The direction is in fact
reversed (CoT slightly outperforms no-CoT by $+2.4$ to $+4.6$\,pp), but
no individual cell reaches Bonferroni significance.
\textbf{Falsified.} H3 was the most directly model-mechanism-tied of the
four hypotheses: it held that the bandwidth bypass would
\emph{actively hurt} TC$^0$ benchmarks because additional serial tokens
add noise without unlocking computation the model lacks bandwidth for.
The data does not support this hypothesis; the bandwidth-bypass
mechanism appears to be one-sided (helping high-depth tasks) rather
than two-sided.

\textbf{H4 (GSM8K largest positive gap):}
Bootstrap 95\% CIs on the CoT recovery gap overlap between GSM8K and MATH
for all models (e.g., Qwen-7B: GSM8K [64.4, 71.5]\,pp vs MATH [62.7, 70.1]\,pp).
GSM8K has the largest average gap (+60.6\,pp vs MATH +60.5\,pp) but the two
benchmarks are statistically indistinguishable -- consistent with both being
deep $k$-composition tasks of similar assigned depth.
HumanEval/Qwen-32B was originally reported as exceeding both in an earlier
draft, but correct parsing puts it at $\Delta = +23.2$\,pp.
\textbf{Not distinguishable from MATH; see \S\ref{sec:discussion}.}

\paragraph{Cohen's $\kappa$ (pre-registered deviation).}
Pooled Cohen's $\kappa$ between the per-benchmark heuristic primitive labels
and Gemma-2-27B-it's per-item primitives is
$\kappa = 0.293$, well below the pre-registered target $\kappa \geq 0.7$.
Per-benchmark agreement: GSM8K 96\%, MATH 87\%, MMLU 59\%, ARC 1.5\%,
HumanEval 4.3\%.
The judge's recorded reasoning (released alongside the labels) shows the
disagreement is principled rather than noisy: it labels most ARC items as
\texttt{set\_disjointness} (factual recall) rather than \texttt{sparse\_parity},
reflecting that ARC items are answerable by retrieval; it labels most
HumanEval items as \texttt{k\_composition} (sequential function build-up)
rather than \texttt{pointer\_chasing}, reflecting that the canonical
solutions of most HumanEval problems chain operations rather than dereference
references through state.
We interpret this as evidence that real benchmarks are mixtures of CC
primitives rather than instances of a single primitive, consistent with
recent work on benchmark heterogeneity.
The depth-based hypotheses tested in H1--H4 are independent of primitive
labels: the failure pattern depends on per-item serial depth, not on which
primitive a benchmark exemplifies. We retain the heuristic primitive labels
as a coarse taxonomy (cf.\ Table~\ref{tab:mapping}) but report the $\kappa$ deviation
transparently. The full judge labels and reasoning text are included in the
released artefact for re-analysis.

\subsection{Per-benchmark Depth Gradient}

We stratify each benchmark by per-item CC depth ($k{=}1$, 2--3, 4--6, $\geq{7}$).
The depth proxy is benchmark-specific: equation count for MATH, calculator-step
count for GSM8K, AST nesting depth for HumanEval; MMLU and ARC items collapse
to $k{=}1$ under any structural depth metric and are excluded from the
gradient analysis (see \S\ref{sec:limitations}).
Table~\ref{tab:kbin} shows the same monotone pattern across all three
benchmarks for which depth is defined.
The Qwen-32B MATH cell illustrates the canonical pattern:
no-CoT accuracy degrades from 45.5\% at $k{=}1$ to 15.4\% at $k{\geq}7$
(a 30\,pp range), while CoT stays in the 82--88\% band -- externalising
computation compensates for depth regardless of $k$. The GSM8K and HumanEval
panels show the same direction of effect with smaller dynamic range.

As a pre-registered secondary analysis, mean CoT output length (completion
tokens) increases monotonically with bin on MATH: 558 tokens at $k$=1, 586 at
$k$=2--3, 662 at $k$=4--6, and 839 at $k$$\geq$7.
A per-item Spearman rank correlation between $k$ and completion-token count
yields $\rho = 0.221$, $p < 10^{-26}$ ($n=2{,}317$ items), confirming that the
model allocates more computation to deeper problems even within MATH.

\begin{table}[ht]
\centering
\small
\setlength{\tabcolsep}{4pt}
\begin{tabular}{llrrrr}
\toprule
\textbf{Bench} & \textbf{Cond.} & $k{=}1$ & $k{=}2{-}3$ & $k{=}4{-}6$ & $k{\geq}7$ \\
\midrule
\multirow{2}{*}{MATH (Qwen-32B)}
 & no-CoT & 45.5 & 31.0 & 21.7 & 15.4 \\
 & CoT    & 88.3 & 88.4 & 85.1 & 82.7 \\
\midrule
\multirow{2}{*}{GSM8K (Qwen-32B)}
 & no-CoT & 54.2 & 53.8 & 18.0 &  7.1 \\
 & CoT    & 97.9 & 94.8 & 91.5 & 92.9 \\
\midrule
\multirow{2}{*}{HumanEval (Qwen-32B)}
 & no-CoT & 28.6 & 17.7 &  0.0 & --- \\
 & CoT    & 95.2 & 83.2 & 86.2 & --- \\
\bottomrule
\end{tabular}
\caption{Accuracy (\%) by CC depth bin for Qwen-32B across the three benchmarks
with non-trivial depth diversity. No-CoT degrades monotonically with $k$ on all
three; CoT is approximately bin-invariant. HumanEval $k{\geq}7$ has $n{=}1$
and is omitted. Bin counts: MATH $n{=}231$/336/175/52; GSM8K $n{=}48$/444/294/14;
HumanEval $n{=}21$/113/29/1.}
\label{tab:kbin}
\end{table}

\subsection{Model Size and No-CoT Accuracy}

\begin{figure}[ht]
    \centering
    \includegraphics[width=0.92\textwidth]{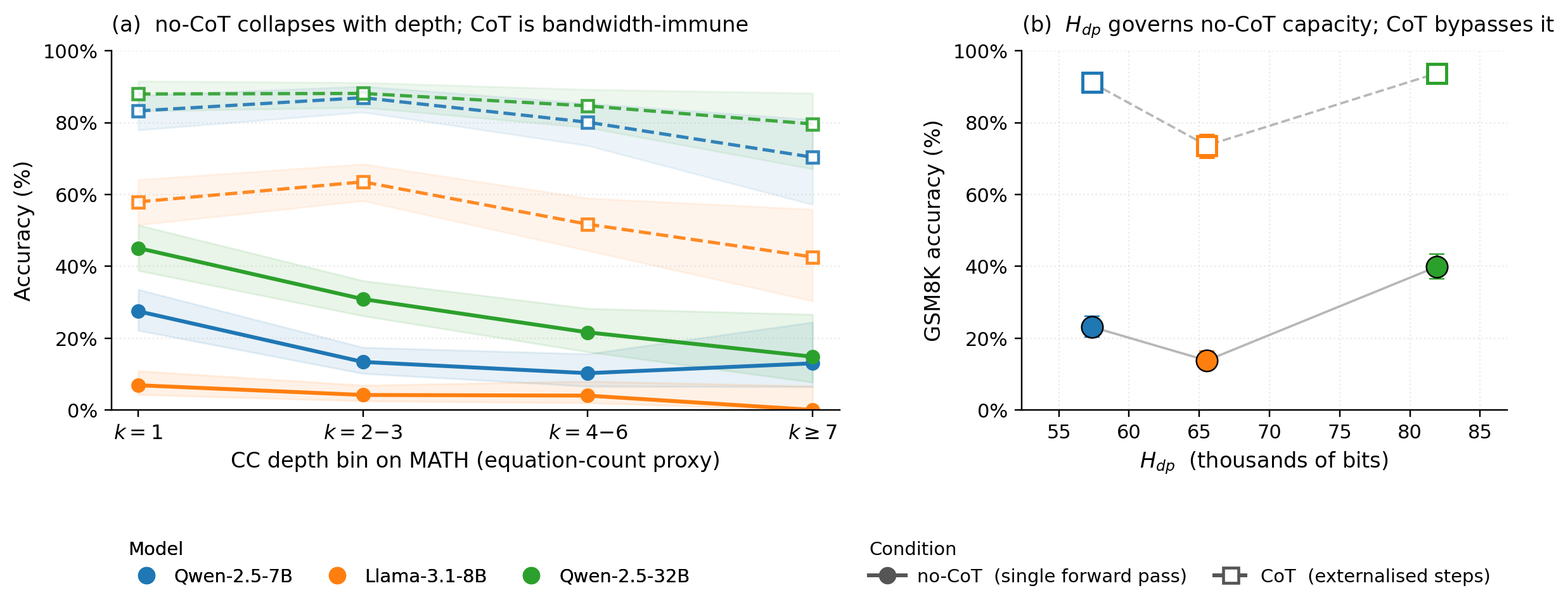}
    \caption{\textbf{No-CoT accuracy, serial depth, and model size.}
    \textbf{(a)}~MATH accuracy stratified by CC depth bin for all three models
    (95\% Wilson bands from per-item data, $n{=}324$--$2{,}022$ per bin).
    No-CoT (solid) collapses monotonically as serial depth grows; CoT (dashed)
    is approximately depth-invariant across the entire range, separating the two
    regimes by 50--75\,pp at the deepest bin.
    \textbf{(b)}~On GSM8K, no-CoT accuracy rises with model size within the Qwen
    family (Llama-8B is off-trend, plausibly because GQA reduces its effective
    per-token bandwidth), while CoT recovers a near-uniform 70--94\% irrespective
    of architecture. With three models and a family confound we read panel~(b)
    as suggestive, not as a quantitative law.}
    \label{fig:depth_scaling}
\end{figure}

Within the Qwen family, no-CoT accuracy on GSM8K rises with model size:
23.1\% (Qwen-2.5-7B, $H_{\!dp}$=57K) to 39.9\% (Qwen-2.5-32B, $H_{\!dp}$=82K),
qualitatively consistent with larger models performing more serial computation
within a single pass. The cross-family comparison is confounded: Llama-3.1-8B
(nominal $H_{\!dp}$=65K) scores only 13.8\%, plausibly because Grouped-Query
Attention (8 KV heads against 32 query heads) reduces its effective per-token
bandwidth. With only three models and a family confound, we do not claim that
$H_{\!dp}$ quantitatively predicts accuracy; we report the trend as suggestive
and qualitatively in line with the serial-depth account.
Figure~\ref{fig:depth_scaling} shows the depth-gradient effect (panel~a)
and the model-size trend (panel~b).

\subsection{Memorization Separation (GSM-Symbolic)}
\label{sec:memo}

An alternative reading of the GSM8K gap is item-level memorization: models
recall pretraining-seen answers under no-CoT, and CoT plays no architectural
role. We test this by re-running all three models on GSM-Symbolic
\citep{mirzadeh2024gsm}, which substitutes names and numerals in GSM8K
templates while preserving arithmetic depth, holding the protocol fixed
(800 items per cell, both conditions, greedy decoding).

Per-cell deltas stay within $\pm 10$\,pp (mean $-4.3$\,pp no-CoT,
$-1.4$\,pp CoT; Table~\ref{tab:memo}) -- well within the GSM8K$\to$GSM-Symbolic
drops reported in \citet{mirzadeh2024gsm} (their Fig.~3 spans $-0.3$ to
$-9.2$\,pp). The CoT recovery gap is preserved in direction and magnitude
on every model: Qwen-7B $+68.0\to+65.4$, Llama-8B $+59.9\to+65.4$,
Qwen-32B $+53.9\to+59.8$\,pp. Memorization would flatten the gap under
perturbation; it does not. The depth-dependent pattern is therefore an
architectural effect, not retrieval of memorized items.

\begin{table}[ht]
\centering
\small
\begin{tabular}{lcccc}
\toprule
Model & Condition & GSM8K (\%) & GSM-Symbolic (\%) & $\Delta$ (pp) \\
\midrule
Qwen-7B   & no-CoT & 23.1\,[20.3,26.2] & 19.2\,[16.7,22.1] & $-3.9$ \\
Qwen-7B   & CoT    & 91.1\,[89.0,92.9] & 84.6\,[82.0,87.0] & $-6.5$ \\
Llama-8B  & no-CoT & 13.8\,[11.5,16.3] & 14.1\,[11.9,16.7] & $+0.4$ \\
Llama-8B  & CoT    & 73.6\,[70.5,76.6] & 79.5\,[76.6,82.2] & $+5.9$ \\
Qwen-32B  & no-CoT & 39.9\,[36.5,43.3] & 30.4\,[27.3,33.6] & $-9.5$ \\
Qwen-32B  & CoT    & 93.8\,[91.9,95.2] & 90.1\,[87.9,92.0] & $-3.6$ \\
\bottomrule
\end{tabular}
\caption{\textbf{GSM8K vs.\ GSM-Symbolic accuracy ($n{=}800$ items per cell,
Wilson 95\% CIs).} All six per-cell deltas fall within $\pm 10$\,pp; the
CoT recovery gap on the math side is preserved on every model. The
positive (math-side) hypothesis of the $H_{\!dp}$ framework is not an
artefact of GSM8K memorisation.}
\label{tab:memo}
\end{table}

\section{Discussion}
\label{sec:discussion}

\textbf{The math-side hypothesis holds; the TC$^0$ hypothesis does not.}
Our pre-registered hypotheses posited a two-sided effect: CoT should help on
high-depth tasks and hurt on low-depth tasks. With correctly extracted
answers, only the first half is borne out. P-complete benchmarks
(GSM8K, MATH) recover $+54$ to $+68$\,pp across all three models,
significant beyond any reasonable correction. TC$^0$ benchmarks (MMLU,
ARC) are essentially flat: CoT changes accuracy by $0.0$ to $+4.6$\,pp
across all six (model, benchmark) cells. The framework's positive
hypothesis (CoT bypasses bandwidth on high-depth tasks) is supported;
the negative hypothesis (additional serial tokens \emph{hurt} TC$^0$
tasks) is not. This narrows the framework's empirical content to a
one-sided claim and converges with the large-scale meta-analysis of
\citet{sprague2025cot}, who report that CoT helps mainly on math and
symbolic reasoning across 14 models and 20 datasets, with little to no
effect on non-symbolic tasks.

\textbf{Why the TC$^0$ hypothesis may have failed.}
Several non-exclusive explanations are consistent with the data.
\emph{Ceiling effects:} no-CoT accuracy on ARC is 82--95\% across the three
models. Any negative CoT effect must come out of a small remaining
headroom; the bandwidth-bypass mechanism would need to specifically
disrupt ceiling-level performance, which is a stronger claim than the
positive (recovery) version. \emph{Instruction-tuning produces a
two-mode model:} our models are RLHF-tuned to follow instructions for
both ``answer directly'' and ``think step by step''. A direct-answer
prompt under a 32-token cap and a CoT prompt under a 2048-token cap may
both produce competent answers for items that fit in a single forward
pass, with neither mode forcing the model into a regime where serial
expansion is harmful. \emph{Mixture-of-primitives:} the pre-registered
LLM-as-judge analysis (\S\ref{sec:results}, $\kappa = 0.293$) already
indicated that real benchmarks are mixtures rather than pure instances
of a single CC primitive. If individual MMLU and ARC items vary
internally in computational depth, the aggregate CoT effect averages
across items where CoT helps (deep) and items where CoT is neutral
(shallow), with the net at approximately zero.

\textbf{Alternative explanations: artefacts and contamination.}
\citet{gururangan2018annotation} show that NLI benchmarks
contain surface-level artifacts exploitable by hypothesis-only classifiers
(67\% accuracy on SNLI without the premise).
An analogous effect may operate here: the high no-CoT accuracy on MMLU (68--82\%)
and ARC (82--95\%) could partly reflect models exploiting lexical shortcuts rather
than performing genuine set-disjointness or sparse parity computation.

A third possibility is data contamination.
\citet{magar2022data} show that pretrained models can
\emph{exploit} (not merely memorize) test labels seen during pretraining,
with exploitation increasing with duplication frequency and model size.
MMLU and ARC are among the most widely distributed NLP benchmarks and are
almost certainly present in the pretraining corpora of all three models tested.
The scale of the no-CoT accuracy on ARC is striking:
\citet{clark2018thinksolvedquestionanswering} report that in 2018 no system
significantly outperformed random ($\approx$25\%) on the ARC Challenge Set --
a partition specifically designed to resist surface-level methods -- yet our
models achieve 82--95\% under no-CoT on the same questions.
A more than triple accuracy jump over the strongest 2018 baselines on
a dataset designed to be hard for retrieval and co-occurrence is
consistent with -- but does not by itself demonstrate -- pretraining
exposure. Modern models also reflect much larger pretraining corpora,
better architectures, and instruction tuning, any of which could
account for substantial portions of the gap.
\citet{hendrycks2021measuring} conducted a memorisation check in 2021:
prompt-compression entropy was not positively correlated with accuracy for
GPT-3, suggesting exact question memorisation was minimal at that time.
However, that analysis predates the instruction-tuning era; our models were
trained on corpora assembled in 2024--2025, in which MMLU has become a standard
evaluation target ubiquitous in model-card reports and leaderboards, making
exposure far more likely than in 2021.
Notably, unspecialized crowd workers (MTurk) achieve only 34.5\% on MMLU
while expert-level accuracy is $\approx$89.8\%~\citep{hendrycks2021measuring};
our models' 68--82\% no-CoT accuracy places them well above unspecialized humans,
suggesting genuine knowledge acquisition from pretraining -- but does not rule
out additional exploitation of memorised label associations for specific subjects.

We consider the annotation-artifact and contamination accounts
\emph{complementary} to the architectural one rather than competing: both
help explain why no-CoT accuracy on TC$^0$ benchmarks is already so close
to ceiling (82--95\% on ARC), which in turn leaves very little headroom
for any architectural effect (positive or negative) to manifest at
the aggregate level. Disentangling these accounts requires benchmarks
explicitly designed to resist lexical shortcuts and verified to be absent
from pretraining corpora.
We leave this to future work.

\textbf{The one significant CoT penalty (HumanEval/Qwen-7B).}
The single Bonferroni-significant negative effect in our data is Qwen-7B on
HumanEval ($-28.7$\,pp, $p = 4\times10^{-9}$): chain-of-thought \emph{lowers}
the smallest model's code accuracy. We flag this as a genuine but unexplained
finding rather than fold it into the serial-depth account, for two reasons.
First, the no-CoT HumanEval condition permits up to 256 tokens to emit a
function body, so it is not a clean single-pass contrast; it is better read as
``write code directly'' versus ``explain first, then write code,'' which the
bandwidth framing does not address. Second, the penalty appears only for the
weakest model, consistent with the broader finding that CoT is not universally
beneficial and can reduce accuracy in some settings \citep{sprague2025cot}.
A mechanistic account of why externalised reasoning harms smaller code models
is left to future work.

\textbf{H4 deviation.}
GSM8K and MATH have statistically indistinguishable CoT recovery gaps.
Both are $k$-composition tasks; their similar gaps are consistent with theory
and strengthen H2 rather than undermining it.

\textbf{Limitations.}
\label{sec:limitations}
\emph{Instruction-tuning confound:} All models are instruction-tuned variants
trained to produce CoT-style reasoning. The short no-CoT caps suppress this
trained behaviour. The math-side recovery effect within the same models
is therefore established under conditions that genuinely deny the model
the externalisation channel; whether the absence of a TC$^0$ penalty
would survive a more aggressive denial of CoT-style reasoning under
no-CoT (e.g.\ explicit ``answer in one token'' instructions) is
untested.
\emph{Test-time compute confound:} The no-CoT (short-cap) and CoT (2048-token)
conditions differ in both whether intermediate state is externalised \emph{and}
the total inference-time compute available. Because a decoder-only transformer
can perform additional serial computation only by emitting tokens, these two
factors are intrinsically entangled, and our design does not separate them; the
results are equally consistent with an inference-time compute-scaling account,
and we do not claim a mechanism distinct from externalised test-time
computation.
\emph{TC$^0$ baseline validity:} No-CoT accuracy on MMLU/ARC is high (68--95\%),
consistent with pretraining contamination or lexical shortcuts
(\S\ref{sec:discussion}). We therefore cannot attribute the absence of a CoT
effect on these benchmarks to the architectural account specifically; the
TC$^0$ null is confounded with ceiling and contamination effects.
\emph{Asymptotic bound:} $H_{\!dp}$ is an asymptotic lower bound;
threshold values are treated as ordinal, not exact.
\emph{Scope:} Decoder-only dense transformers only; MoE and SSM architectures
are excluded.
\emph{English-only benchmarks.}
\emph{Heuristic depth labelling:} The per-item depth estimator (calculator-step
count for GSM8K, equation-count proxy for MATH, AST nesting depth for HumanEval)
produces non-trivial bin diversity for those three benchmarks. MMLU and ARC
items collapse to $k{=}1$ under any structural depth metric we tried, because
their per-item textual structure does not expose a numeric depth proxy. We
treat this collapse as consistent with their TC$^0$ classification rather than
as a measurement artefact, but bin-stratified analyses for MMLU and ARC are
correspondingly limited to the $k{=}1$ stratum.
\emph{Theoretical framing (pre-registered deviation):} The pre-registration
framed H1--H4 as predictions of the $H_{\!dp}$ bound. On closer analysis the
bound is asymptotic: its failure guarantee is non-vacuous only for prompt
lengths $n^\star = H_{\!dp}^{\,2^{4L}}$ (here $\log_{10} n^\star \gtrsim 10^{34}$),
so it does not bind at the context lengths we test. We therefore reframe it as
conceptual motivation rather than a literal predictor. The hypotheses and all
registered analyses are unchanged from registration; only their theoretical
interpretation is refined.
\emph{Cohen's $\kappa$ deviation:} The pre-registered LLM-as-judge inter-rater
check (described in the Scoring section) yielded $\kappa = 0.293$, well below
the pre-registered $\kappa \geq 0.7$ target. The judge's per-item labels
indicate that the heuristic single-primitive-per-benchmark labels are
too coarse: ARC items are
predominantly judged \texttt{set\_disjointness} rather than the pre-registered
\texttt{sparse\_parity}, and HumanEval items are predominantly judged
\texttt{k\_composition} rather than \texttt{pointer\_chasing}. We retain
the heuristic primitive taxonomy as a coarse descriptor (cf.\ Table~\ref{tab:mapping}) and
report this as a pre-registered deviation. The depth-based hypotheses tested
in H1--H4 are not affected by primitive labels.
\emph{Pre-registration completion:} The memorization-separation analysis,
originally deferred, was completed via GSM-Symbolic perturbations
\citep{mirzadeh2024gsm} and is reported in
Section~\ref{sec:memo}. All other pre-registered analyses (functional
HumanEval scoring with the full 984-row execution sweep; the bin-level
McNemar tests on real per-item depth labels for GSM8K/HumanEval; the
LLM-as-judge $\kappa$ check) were completed for this version of the paper. All main-text numerical claims are recomputed by \texttt{verify\_paper\_stats.py};
Section~\ref{sec:memo} claims are reproduced by scripts in the
\texttt{memorization\_check/} directory, released as a separate archive.

\section{Related Work}

\textbf{Transformer expressivity.}
\citet{hahn2020theoretical} proved (Corollary~2 therein) that hard-attention
transformers cannot compute PARITY or model 2DYCK, and extends these limitations
to soft attention under smoothness assumptions on activations.
The PARITY result is direct theoretical precedent for our ARC-Challenge
``sparse parity'' primitive classification: transformers are provably unable to
compute parity regardless of scale, which is consistent with both the high no-CoT baseline
and the absence of CoT recovery we observe on ARC.
\citet{perez2021attention} proved Turing-completeness for
the Transformer under hard attention and \emph{arbitrary-precision} activations,
but explicitly note that \textit{``Transformers with fixed precision are not
Turing complete.''}
Since our models operate at FP16 (fixed precision), they are precisely in the
non-Turing-complete regime -- creating the theoretical space for serial-depth
limitations to manifest.
\citet{chen2024theoreticallimitationsmultilayertransformer} close this
gap by proving the first unconditional lower bounds for multi-layer decoder-only
transformers via the autoregressive communication model, with $H_{\!dp}$ as the
governing bandwidth parameter under realistic (FP16) precision.

\textbf{Chain-of-thought.}
\citet{wei2022cot} established CoT as a general reasoning enhancement,
and \citet{wei2022emergent} showed CoT is itself an \emph{emergent ability}
that only surpasses direct-answer prompting on GSM8K at ${\sim}68$B parameters
(base LaMDA; Figure~3A therein).
Our instruction-tuned 8B model already gains $+59.9$\,pp from CoT, consistent
with RLHF shifting the emergence threshold downward.
Critically, \citet{wei2022emergent} frame CoT as uniformly beneficial
once it emerges; our results show its benefit is concentrated on
P-complete (math) tasks and is approximately zero on TC$^0$ tasks
(MMLU, ARC) across all three models tested -- a distinction their
scaling analysis does not capture.
\citet{cobbe2021trainingverifierssolvemath} show that finetuning a
6B model to answer GSM8K \emph{without} intermediate steps collapses accuracy
from 20.6\% to 5.2\%, with the authors explicitly attributing this to the model
having no mechanism to route intermediate computation outside a single forward
pass -- the same bottleneck formalised by the $H_{\!dp}$ bound.
\citet{nye2021scratchpad} introduced the scratchpad, showing that
polynomial evaluation improves from 8.8\% to 20.1\% (few-shot) when a model
writes intermediate steps rather than answering directly, with even larger
gains for addition on out-of-distribution lengths.
Their core mechanistic observation -- that the model \textit{``cannot adapt
the amount of compute''} within a single forward pass -- is the empirical
precursor to the $H_{\!dp}$ bound.
Crucially, Nye et al.\ find scratchpad helps most on tasks requiring
sequential multi-step computation, directly foreshadowing our result that
only P-complete benchmarks benefit from externalising computation, while
TC$^0$ tasks do not.
In the code-generation domain, \citet{chen2021evaluatinglargelanguagemodels} provide
independent empirical evidence of the same bottleneck: Codex's pass rate on
synthetic problems built from chained string-manipulation primitives drops by
a factor of 2--3 with every additional chained operation (their Figure~11),
and the model fails to bind operations to variables once chain length exceeds
a few steps.
These findings predate the $H_{\!dp}$ bound and are consistent with it: single
forward-pass bandwidth limits serial computation regardless of domain.
Subsequent work~\citep{kojima2022zeroshot, wang2023selfconsistency} extended CoT
to zero-shot and ensemble settings but likewise treats it as a universal gain.
\citet{sprague2025cot} conduct the most comprehensive empirical
challenge to this universality assumption: a meta-analysis of 100+ papers and
experiments across 20 datasets and 14 models show that CoT helps mainly on
math and symbolic reasoning, with negligible or negative effects on commonsense,
knowledge, and soft-reasoning tasks.
They further decompose CoT's benefit into \emph{planning} (translating a problem
into a formal specification) and \emph{execution} (performing intermediate
symbolic steps), showing that much of CoT's gain comes from execution -- where
it nonetheless underperforms external symbolic solvers.
Our results converge with theirs on the positive side: CoT's benefit on
math is supported by both the $H_{\!dp}$ bandwidth-bypass account and
their planning/execution decomposition. On the negative side, our
corrected data show that the absence of a CoT benefit on TC$^0$ tasks
is not accompanied by a systematic CoT penalty, which constrains the form of any account
that would attribute the null to the bound itself rather than to
ceiling effects, instruction-tuning, or task heterogeneity.

\textbf{Benchmark analysis.}
\citet{gururangan2018annotation} showed that crowdsourced
NLI benchmarks contain annotation artifacts exploitable without genuine inference
-- hypothesis-only models reach 67\% on SNLI.
\citet{magar2022data} showed that models can
\emph{exploit} (not merely memorize) contaminated test labels seen during
pretraining, with exploitation growing with duplication frequency and model size;
critically, memorization does not guarantee exploitation, implicating
specific training dynamics.
Both confounds are directly relevant to the high no-CoT scores we observe on
TC$^0$ benchmarks (\S\ref{sec:discussion}).
We introduce CC primitive labeling as a complementary lens for anticipating
\emph{which} benchmarks benefit from which architectural capabilities,
orthogonal to both artifact and contamination analyses.

\section{Conclusion}

We investigate whether the serial-depth bottleneck identified by the
$H_{\!dp}$ bandwidth bound \citep{chen2024theoreticallimitationsmultilayertransformer}
governs behaviour on standard NLP benchmarks, even though the formal bound binds
only at astronomically large context lengths ($n^\star = H_{\!dp}^{\,2^{4L}}$).
Across three instruction-tuned models and five
benchmarks, the framework's positive hypothesis is supported: P-complete
tasks (GSM8K, MATH) show $+54$ to $+68$\,pp CoT recovery gaps across all
three models, and HumanEval shows the hypothesised model-size-dependent
transition ($+23.2$\,pp for Qwen-32B; $-28.7$\,pp for Qwen-7B). The
framework's stronger negative hypothesis -- that CoT should
\emph{actively hurt} TC$^0$ tasks -- is not supported: with correctly
extracted answers, CoT is approximately neutral on MMLU and ARC across
all six (model, benchmark) cells ($\Delta \in [0.0, +4.6]$\,pp). The
pooled cross-benchmark depth--recovery correlation is Spearman $\rho = 0.661$ ($p = 0.007$,
$n = 15$); 9 of 15 pre-registered McNemar tests are significant;
pre-registered H3 (MMLU no-CoT $\geq$ CoT) is falsified.

All data, code, judge labels, and inference logs are released to
support independent verification. The $H_{\!dp}$ framework is a useful
one-sided account of where CoT will help; whether it can also anticipate
where CoT actively harms requires benchmarks designed to resist lexical
shortcuts and verified to be absent from pretraining corpora.

\subsubsection*{AI Usage Disclosure}
Claude Opus 4.7 (Anthropic) was used during the preparation of this
manuscript for code generation, literature search, and editing of draft
text. All scientific claims, experimental design, data collection,
analysis, and conclusions are the sole responsibility of the author.
No AI-generated content was used as a primary source or cited as
evidence.

\bibliography{refs}
\bibliographystyle{tmlr}

\appendix
\section{Released Artifacts}
\label{app:data}

The OSF project (\url{https://osf.io/hteuj},
DOI~\texttt{10.17605/OSF.IO/92JDK}) and the Zenodo deposit
(DOI~\texttt{10.5281/zenodo.20294033}) include:
the 20{,}184-record SQLite inference database (31\,MB; raw outputs,
extracted answers, gold labels, correctness flags, and completion-token
counts); all system/user prompts; the regex, \texttt{\textbackslash boxed\{\}},
and HumanEval functional-execution scorers; per-item CC depth labels
(GSM8K calculator steps, MATH equation counts, HumanEval AST depths);
the Gemma-2-27B-it judge labels and reasoning text (964 items);
\texttt{verify\_paper\_stats.py}, which reproduces every main-text
number in under 30\,seconds; \texttt{rescore\_mmlu\_arc.py}
(Appendix~\ref{sec:correction}) and \texttt{verify\_parser\_bug.py},
the rescoring script and the diagnostic that caught the v1 parser
artefact; the \texttt{memorization\_check/} scripts reproducing
\S\ref{sec:memo}; the pre-registration document; and figure-generation
code for Figures~\ref{fig:theorem}--\ref{fig:depth_scaling}.

\section{Scoring Artefact Corrections}
\label{sec:correction}

This is a corrected manuscript that addresses two independent scoring artefacts
present in an early preprint draft (released on Zenodo on 19~May 2026). Both
artefacts artificially suppressed no-CoT baselines.

\paragraph{MMLU and ARC (First Correction).}
The initial draft reported large negative CoT effects on MMLU and ARC-Challenge,
framed as a ``CoT Sign Reversal.'' The MMLU/ARC scorer used the regex
\texttt{\textbackslash b([A-D])\textbackslash b} and returned the \emph{first} standalone capital letter found.
For CoT outputs that enumerate options, this regex reliably returns
\texttt{A} regardless of the model's final answer. A final-answer-aware parser
locates the last ``Answer'' marker, and extracts the case-sensitive letter.
With proper scoring, CoT is approximately neutral on MMLU and ARC across all six cells.

\paragraph{HumanEval (Second Correction).}
An intermediate draft reported a very low no-CoT baseline (15.9\%) for Qwen-32B on HumanEval.
A subsequent audit revealed that the functional execution script
(\texttt{score\_humaneval.py}) did not strip stop tokens (e.g.\ \texttt{<|assistant|>})
from the raw generated text. During functional execution, these tags caused
\texttt{SyntaxError} tracebacks for otherwise correct code, artificially suppressing
the no-CoT accuracy. With tags properly stripped via regex, the true Qwen-32B
no-CoT baseline is 62.2\%, and the CoT benefit is $+23.2$\,pp.

\paragraph{What the corrections change.}
The central ``CoT Sign Reversal'' framing of the initial draft on MMLU and ARC is not supported by the
corrected data. The math-side findings (GSM8K and MATH
CoT recovery gaps of $+54$ to $+68$\,pp) and the GSM-Symbolic memorisation
control are unchanged. The HumanEval model-size-dependent crossover is preserved,
but with smaller effect magnitudes ($+23.2$\,pp for Qwen-32B, $-28.7$\,pp for Qwen-7B).
The pre-registered H1 (originally 15/15 significant)
becomes 9/15; H3 (MMLU no-CoT $\geq$ CoT) is falsified; the
depth--recovery Spearman correlation reduces from $\rho = 0.850$ to
$\rho = 0.661$ (still significant at $p = 0.007$).

\end{document}